\documentclass{article}
\usepackage{ijcai23}

\usepackage{times}
\usepackage{soul}
\usepackage{url}
\usepackage[hidelinks]{hyperref}
\usepackage[utf8]{inputenc}
\usepackage[small]{caption}
\usepackage{graphicx}
\usepackage{amsmath}
\usepackage{amsthm}
\usepackage{booktabs}
\usepackage{algorithm}
\usepackage{algorithmic}
\usepackage[switch]{lineno}
\title{FAGH: Accelerating Federated Learning with Approximated Global Hessian}

\author{
Mrinmay Sen$^{1,2}$
\and
A. K. Qin$^1$\and
Krishna Mohan C$^{2}$
\affiliations
$^1$Dept. of Computing Technologies, Swinburne University of Technology, Hawthorn, VIC, Australia\\
$^2$Dept. of Artificial Intelligence,  Indian Institute of Technology Hyderabad, Hyderabad, India
}%
\begin{document}

\maketitle

\begin{abstract}
In federated learning (FL), the significant communication overhead due to the slow convergence speed of training the global model poses a great challenge. Specifically, a large number of communication rounds are required to achieve the convergence in FL. One potential solution is to employ the Newton-based optimization method for training, known for its quadratic convergence rate. However, the existing Newton-based FL training methods suffer from either memory inefficiency or high computational costs for local clients or the server. To address this issue, we propose an FL with approximated global Hessian (FAGH) method to accelerate FL training. FAGH leverages the first moment of the approximated global Hessian and the first moment of the global gradient to train the global model. By harnessing the approximated global Hessian curvature, FAGH accelerates the convergence of global model training, leading to the reduced number of communication rounds and thus the shortened training time. Experimental results verify FAGH's effectiveness in decreasing the number of communication rounds and the time required to achieve the pre-specified objectives of the global model performance in terms of training and test losses as well as test accuracy. Notably, FAGH outperforms several state-of-the-art FL training methods.
\end{abstract}

\section{Introduction}

In centralized learning, all data is collected in one place and used to train a machine learning model. Despite its high performance, centralized learning poses risks of privacy leakage and communication overhead when collecting data from different sources or clients. These challenges motivate the transition to federated learning from centralized learning. In the most popular and baseline algorithm of federated learning, FedAvg \cite{McMahanaistats2017}, locally trained models are collected on the server instead of raw data, and the server aggregates all local models to find the global model, which is then sent back to all clients for further training. Sending local models instead of raw data to the server helps overcome data transfer challenges. One communication round of FedAvg involves two communications: sending the global model from the server to all available clients and sharing locally trained models with the server, resulting in a communication cost of O(2d). Since FedAvg employs a first-order stochastic gradient descent optimizer \cite{ketkar2017stochastic} to update local models, it is computationally efficient, with a local time complexity of O(d), where d represents the number of model parameters. FedAvg performs well when data are homogeneously distributed across all clients \cite{Liiclr2020}, a scenario which is rarely encountered in real-life applications. In cases of heterogeneous data distribution, FedAvg suffers from objective inconsistency \cite{Karimireddyicml2020,Limlsys2020,Liiclr2020,Tancorr2021,Wangneurips2020}. This inconsistency occurs when the global model converges to a stationary point that may not be the optimum of the global objective function, resulting in slow training of the global model, increased number of communication rounds, and time required to achieve a certain performance level. To accelerate FL training in scenarios of heterogeneous data distribution, several modifications have been proposed for FedAvg, including FedProx \cite{Limlsys2020}, FedNova \cite{Wangneurips2020}, SCAFFOLD \cite{Karimireddyicml2020}, MOON \cite{Licvpr2021}, FedDC \cite{Gaocvf2022}, pFedMe \cite{DinhTN20NeurIPS}, FedGA \cite{DandiBJgradalign2022}, FedExP \cite{fedexp}, among others. Although these modifications generally outperform FedAvg, the learning of the global model remains slower when aiming for a targeted performance, as these methods primarily utilize first-order gradient information for optimizing model parameters. Additionally, these methods are highly sensitive to hyperparameter choices.

To further accelerate FL training, researchers have shifted their attention from first-order optimization to the second-order Newton method of optimization due to its higher convergence rate compared to first-order methods \cite{Agarwaljmlr2017,Tankaria2021corr}. Although the Newton method of optimization outperforms first-order optimizations in term of convergence, there are challenges in calculating and storing the Hessian and its inverse for large-scale settings (with time complexities of $O(d^2)$ and $O(d^3)$ respectively for calculating the Hessian and its inverse, and a space complexity of $O(d^2)$ for storing them). To address these challenges, researchers have focused on approximating the Hessian \cite{Agarwaljmlr2017,Liu1989mp,Martens2015jmlr,Tankaria2021corr,nazareth2009conjugate,vuchkov2022hessian} instead of using the true Hessian while optimizing model parameters. In federated learning, another issue arises when local models are updated using the Newton method of optimization. Since the Newton method of optimization utilizes the Hessian inverse for updating model parameters, averaging all locally trained models to find the global model is not feasible \cite{DDerezinskiM19}. State-of-the-art solutions to these problems can be categorized into three approaches. The first approach involves computing the local Hessian matrix and sending it to the server in a compressed form, where the server aggregates all local Hessian information to determine the global Newton direction \cite{SafaryanIQR22,QianISR22}. The second approach is based on finding the local Newton direction with global gradient information \cite{Wangneurips2018,Dinhieeetrans2022}. The third approach utilizes the Quasi-Newton method of optimization \cite{Macorr2022}, where local first-order information is collected and aggregated on the server to find the global Newton direction. Existing Newton method-based federated optimizations \cite{bischoff2021second} include DANE \cite{Shamiricml2014}, GIANT \cite{Wangneurips2018}, FedDANE \cite{Licorr2020}, FedSSO \cite{Macorr2022}, FedNL \cite{SafaryanIQR22}, Basis Matters \cite{QianISR22}, DONE \cite{Dinhieeetrans2022}, etc., which are either computationally expensive for local clients due to the calculation of the Hessian matrix, memory inefficient due to storing the Hessian matrix, or associated with four times communication in each FL communication round.
 
To expedite FL training, this paper introduces FAGH, a Newton method-based federated learning algorithm that eliminates the need for four separate communications, as seen in approaches like DONE or GIANT. FAGH approximates the global Newton direction on the server without computing and storing the full Hessian matrix. In FAGH, the server collects gradients and the first row of the true Hessian from each local client. Utilizing the first moment of the average gradient and the first moment of the average first row of the true Hessian across all clients, the server determines the global Newton direction and updates the global model. By leveraging the approximated global Hessian curvature, FAGH accelerates the convergence of global model training, resulting in a reduced number of communication rounds and shorter training times. Experimental results confirm FAGH's effectiveness in reducing the required number of communication rounds to achieve predefined objectives for global model performance, including training and test losses, as well as test accuracy. Notably, FAGH outperforms several state-of-the-art FL training methods. Since FAGH utilizes only the first row of the true Hessian when determining the Newton direction, it can significantly reduce local time and space complexities compared to existing second-order FL algorithms.  

The main contributions of FAGH are as follows.
\begin{itemize}
  \item In FAGH, each client finds gradient and Hessian's first row of the local loss function and sends these to the server. Server finds the first moments of global gradient and global Hessian's first row.  
  \item FAGH directly finds global Newton direction with the help of these first moments of global gradient and first row of the global Hessian without storing and calculating the full global Hessian matrix in the server.
  \item Use of this directly computed global Newton direction leads to faster training in federated learning with linear time computational and space complexities.
\end{itemize}
The rest of the paper are arranged as follows. Section 2 elaborates related works, Section 3 elaborates the problem formulation, section 4 gives basic Preliminaries, section 5 discusses about our proposed method, section 6 shows the experimental setup and results and section 7 concludes our whole works.

\section{Related Work}
The related works can be classified into two categories: first-order based and second-order based FL approaches. 

Existings first-order based approaches consist of FedProx, FedNova, SCAFFOLD, MOON, FedDC, pFedMe, FedGA, FedExP etc. In FedProx, the local objective function or loss function is modified by incorporating a proximal term($\mu$), which helps to control the direction of local gradient. Instead of using simple average or weighted average in server , FedNova uses normalized averaging to get the global model. To control drastically fluctuation of  local gradient, which is caused by data heterogeneity, SCAFFOLD uses variance reduction while updating local model. To correct the local training, MOON conducts model-level contrastive learning where the similarity between model representations is utilized. To control the update of local model, FedDC uses an auxiliary local drift variable. Moreau envelopes regularized loss function is used in pFedMe. FedGA finds the displacement of the local gradient with respect to the global gradient and uses this while initiating local models. To speed up FL training, FedExP adaptively finds the server step size or learning rate by using extrapolation mechanism of Projection Onto Convex Sets (POCS) algorithm.

Existings second-order based approaches include DANE, GIANT, FedDANE, FedSSO, FedNL, Basis Matters, DONE etc. FedSSO utilizes server based Quasi-Newton method on global gradient (average gradient across all the clients) to find the global Newton direction. FedSSO has the same local time complexity as FedAvg (O(d)), but it involves with storing of the full Hessian matrix in the server, which may not be practical in the large scale settings to sustain the server space complexity of $O(d^2)$. DANE, GIANT and FedDANE utilize conjugate gradient method to approximate the Hessian and DONE uses Richardson Iteration. DANE, GIANT, FedDANE and DONE require global gradient communicated by the server while finding local Newton direction, which increases  total time for one FL iteration (One FL iteration of these methods consist of four separate communications- 1. Sending initial model from server to clients 2. Sending local gradients from clients to server 3. Sending average gradient from server to clients and 4. Sending locally updated models from clients to server ). Utilization of conjugate gradient method or Richardson Iteration involves time complexity of O(md+d), where m is number of conjugate gradient or Richardson iterations, which increases local time complexity by m times than FedAvg. FedNL and Basis Matters are associated with finding local Hessian information and sending it to the server in a compressed form.  FedNL and Basis Matters store previous step's Hessian matrix to approximate current step's Hessian. Storing, calculation and compression of local Hessian  results in additional computational and memory load to the local clients.

\section{Problem formulation}
Let $C=\{C_1, C_2, .., C_K\}$ is the set of  participating clients in federated learning, where K is number of clients. $D_i$ is the dataset owned by Client $C_i$. The goal of federated learning is to find the optima of the global objective function $F(w)$ $\forall$ $w \in R^d$ as mentioned in eq.\ref{eq:1}

\begin{equation}
    \label{eq:1}
    \min_w F(w)= \sum^K_{i=1} p_i F_i(w; D_i )
\end{equation}

where, $w$ is model parameters, $F_i(w)= \frac{1}{|D_i|}\sum_{ \xi_j \in D_i}f_j(w; \xi_j )$  is the average loss of Client $C_i$ computed on dataset $D_i$, $f_j$ is local loss for sample $\xi_j \in D_i$ and $\sum_{i=1}^K p_i=1$.

\section{Preliminaries}
\subsection{Newton method of optimization}
Newton method of optimization is similar to first-order stochastic gradient descent (SGD) \cite{ketkar2017stochastic} with only difference in finding update direction. In SGD, the gradient of objective function is scaled by a learning rate parameter $\eta$ to find update direction, which is shown in eq.\ref{eq:2}. But in Newton method (eq. \ref{eq:3}), the update direction is found by scaling the gradient with the help of inverse of the true Hessian (H), which incorporates curvature information while searching for optima of the objective function. Use of second-order Hessian curvature while optimizing model parameters leads to quadratic convergence rate \cite{Agarwaljmlr2017}, which motivates us to use Newton method in FL for accelerating global model training.  

\begin{equation}
    \label{eq:2}
    w_t= w_{t-1} - \eta g_{t-1}
\end{equation}
\begin{equation}
    \label{eq:3}
    w_t= w_{t-1} - H^{-1} g_{t-1}
\end{equation}

\subsection{Sherman Morrison formula of matrix inversion}
The inverse of the the matrix (B + $ZV^T$) $\in R^{d \times d}$ can be calculated using Sherman Morrison formula of matrix inversion as shown in below equation.
\begin{equation}
    (B + ZV^T)^{-1}= B^{-1} - \frac{B^{-1}ZV^TB^{-1}}{1+V^TB^{-1}Z}
\end{equation}

Where, B $\in R^{d \times d}$ be a invertible square matrix and Z, V $\in R^{d \times 1}$ are column vectors.

%\subsection{Chain rule of Leibniz's notation}

\begin{algorithm}[!h]
   \caption{FAGH}
   \label{alg:algo1}
\begin{algorithmic}[1]
    \item \textbf{Input:} $T$: Number of global epochs, $w_0$: Initial global model, $\eta$: learning rate, $\rho$: Hessian regularization parameter, \{$\beta_1, \beta_2\} \in [0, 1)$: Exponential decay rates for the moment estimates, \{$M_1^0 \leftarrow 0, M_2^0 \leftarrow 0$\}:  Initial moment vectors which are initialized with zero  \newline
    \FOR{$t=1$ {\bfseries to} $T$}
        \STATE Randomly pick a subset of clients $\overline{C} \subseteq C $
        \STATE Server sends global model $w_{t-1}$ to all the available clients $\overline{C}$
        \STATE \underline{\textbf{In clients:}}\\
        \FOR{$i=1$ {\bfseries to} $|\overline{C}|$} 
            \STATE Client $C_i$ finds local gradient $g_i^t$= $\frac{\partial F_i(w_{t-1}, D_i)}{\partial w_{t-1}}$ and first row of the true hessian $v_i^t$= $\frac{\partial g_i^t[0]}{\partial w_{t-1}}$, where $g_i^t[0]$ is the first element of $g_i^t$
        \ENDFOR
        \STATE \underline{\textbf{In server:}}\\
        \STATE Collect $g_i^t$ and $v_i^t$ from all the clients
        \STATE Aggregate all $g_i^t$ to find global (average) gradient $g^t=\sum p_i g_i^t$ and aggregate all $v_i^t$ to find first row of the global (average) Hessian $v^t=\sum p_i v_i^t$ across all the clients
        \STATE Find $M_1^t$= $\beta_1 M_1^{t-1} + (1-\beta_1) g^t$
        \STATE Find $M_2^t$= $\beta_2 M_2^{t-1} + (1-\beta_2) v^t$
        \STATE Find $\widehat{M_1}^t= \frac{M_1^t}{1-\beta_1^t}$ and $\widehat{M_2}^t= \frac{M_2^t}{1-\beta_2^t}$,  here $\beta_j^t$ is $\beta_j$ to the power t
        \STATE $G^t\leftarrow$ $\widehat{M_1}^t \in R^{d \times 1}$
        \STATE $V^t\leftarrow$ $\widehat{M_2}^t \in R^{d \times 1}$
        \STATE Find $Z^t$=$\frac{V^t}{V^t[0]} \in R^{d \times 1}$, where $V^t[0]$ is the first element of $V^t$
        \STATE Calculate update $(H_a+ \rho I)^{-1} G^t$=$\frac{G^t}{\rho}$ - $\frac{\frac{Z^t {V^t}^T G^t}{\rho^2}}{1+\frac{{V^t}^T Z^t}{\rho}}$
        \STATE Find $w_{t}$ = $w_{t-1}$ - $\eta (H_a + \rho I)^{-1} G^t$
    \ENDFOR
    %\newline
\end{algorithmic}
\end{algorithm}

\section{Proposed Method}
One communication round of FAGH is shown in algo. \ref{alg:algo1}. In our proposed FAGH, at communication round t, server first sends the global model $w_{t-1}$ to all the available clients. Each client $C_i$ uses this model $w_{t-1}$ as initial model and finds local gradient $g_i^t$ and first row of the true Hessian $v_i^t$ with the help of their local data and local optimizer and shares $g_i^t$ and $v_i^t$ to the server. The server then aggregates all the local gradients to find global (average) gradient $g^t$ across all the clients and aggregates all the local Hessian's first rows to find first row of the global (average) Hessian $v^t$ and finds their first moments (exponential moving averages) $G^t$ and $V^t$ respectively. The server utilizes $V^t$ to approximate the global Hessian $H_a$ and scales $G^t$ directly with the inverse of the regularized $H_a$ using the Sherman-Morrison formula for matrix inversion. Then the server uses this scaled $G^t$ to find the global Newton direction.

\subsection{Hessian approximation with first row of the true Hessian}
Let $w \in R^{d \times 1}= \{x_1, x_2, ..., x_d \}$ be the model parameters to be optimized. We approximate the Hessian with the help of first row of the true Hessian using \textbf{Statement 1}. 

\textbf{Statement 1}  The Hessian of a twice differentiable loss function F with respect to $w$ can be approximated by using eq. \ref{eq:4}. 

\begin{equation}
    \label{eq:4}
    H_a= \frac{VU^T}{\frac{\partial^2 F}{\partial x_1^2}}
\end{equation}
Where, \textbf{V} $\in R^{d \times 1}$ =$\{ \frac{\partial^2F}{\partial x_1^2}, \frac{\partial^2F}{\partial x_1\partial x_2}, \frac{\partial^2F}{\partial x_1\partial x_3},....,\frac{\partial^2F}{\partial x_1\partial x_d}\}$ is first row of the true Hessian ,\\
\textbf{U} $\in R^{d \times 1}$ =$\{\frac{\partial^2F}{\partial x_1^2}, \frac{\partial^2F}{\partial x_2\partial x_1}, \frac{\partial^2F}{\partial x_3\partial x_1},....,\frac{\partial2^F}{\partial x_d\partial x_1}\}$ is first column of the true Hessian. \\
\textbf{Proof of Statement 1:}
 Let H $\in$ $R^{d\times d}$ be the true Hessian of the loss function F. Then the $(i, j)^{th}$ element of H is $H^{(i,j)}$= $\frac{\partial^2 F}{\partial x_i \partial x_j}$, where i,j $\in$ \{1, 2, 3, ...,d\}. 
\\Now, according to eq. \ref{eq:4}, $(i, j)^{th}$ element of approximated Hessian ($H_a$) is found as $H_a^{(i,j)}$ = $\frac{V^{j} \times U^i}{\frac{\partial^2 F}{\partial x_1^2}}$, where $V^j$ = $\frac{\partial^2F}{\partial x_1\partial x_j}$ and $U^i$ = $\frac{\partial^2F}{\partial x_i\partial x_1}$ are $j^{th}$ and $i^{th}$ elements of V and U respectively. So, we can write\\

$H_a^{(i,j)} = \frac{V^{j} \times U^i}{\frac{\partial^2 F}{\partial x_1^2}} = \frac{\frac{\partial^2F}{\partial x_1\partial x_j} \times \frac{\partial^2F}{\partial x_i\partial x_1}}{\frac{\partial^2 F}{\partial x_1^2}}  = \frac{ \frac{\partial(\frac{\partial F}{\partial x_1})}{\partial x_j} \times \frac{\partial( \frac{\partial F}{\partial x_i})}{\partial x_1}}{\frac{\partial (\frac{\partial F}{\partial x_1})}{\partial x_1}} $\\

Using chain rule of Leibniz's notation \cite{swokowski1979calculus}, we rewrite the above expression by assuming that the objective function F is a non-linear function. Let, y is scalar valued output from the model (for classification task, y is considered as the output of corresponding true class for the input sample). If F is a non-linear function, then we can express $\frac{\partial F}{\partial x_i}$ as a function of y for all $i \in \{1, 2,..., d\}$, and we can utilize the chain rule of Leibniz's notation to reformulate the above expression as follows. For example, if $F = \log (y)$, then  $\frac{\partial F}{\partial x_i} = \frac{1}{y} \frac{\partial y}{\partial x_i}$, which is a function of y. Another example: if $F = (y-a)^2$, then  $\frac{\partial F}{\partial x_i} = 2(y-a) \frac{\partial y}{\partial x_i}$, which is also a function of y, here a is any constant. 

$H_a^{(i,j)} = \frac{ \frac{\partial(\frac{\partial F}{\partial x_1})}{\partial y}\times \frac{\partial y}{\partial x_j} \times \frac{\partial( \frac{\partial F}{\partial x_i})}{\partial y} \times \frac{\partial y}{\partial x_1}}{\frac{\partial (\frac{\partial F}{\partial x_1})}{\partial y} \times \frac{\partial y}{\partial x_1}} $\\

$H_a^{(i,j)} = \frac{\partial(\frac{\partial F}{\partial x_i})}{\partial y} \times \frac{\partial y}{\partial x_j} = \frac{\partial^2 F}{\partial x_i \partial x_j}$\\

So, we can write $H_a^{(i,j)}= H^{(i,j)}$\\

Which indicates that we can use eq. \ref{eq:4} for approximation of the Hessian.
  
As Hessian is symmetric matrix,  we can say that the first column and the first row of the Hessian are identical. So we can say U=V. 
Putting U=V in eq. \ref{eq:4} we get,
\begin{equation}
    \label{eq:5}
    H_a= \frac{VV^T}{\frac{\partial^2 F}{\partial x_1^2}} = Z V^T
\end{equation}
So, eq. \ref{eq:5} helps us to approximate the full Hessian with the help of only first row of the true Hessian, where Z= $\frac{V}{\frac{\partial^2 F}{\partial x_1^2}}$. 

\subsection{Finding global Newton direction}
In our proposed method FAGH, server uses global gradient $g_t$ and first row of the global Hessian $v_t$ while finding global Newton direction (where $g_t$ and $v_t$ are found by aggregating local gradients and local Hessian's first rows respectively). Taking inspiration from ADAM \cite{KingmaB14}, we use hyper-parameters $\beta1$, $\beta2$ $\in$ [0, 1) for finding the exponential moving averages $M_1^t$ and $M_2^t$ of $g^t$ and $v^t$ respectively, which helps to update the global model without forgetting knowledge gained in previous communication rounds that is well suited for partial device participation, where all the clients may not be available at a certain communication round. As $M_1^t$ and $M_2^t$ are initialized as (vectors of) 0’s, these are biased towards zero. So we use bias-corrected
estimates $\widehat{M_1}^t$ and $\widehat{M_2}^t$, which have been shown in algo. \ref{alg:algo1}. We find approximated global Hessian ($H_a$) by putting $V^t$ = 
$\widehat{M_2}^t$ in eq. \ref{eq:5} and use the following regularized variant of Newton type update to avoid forming indefinite global Hessian
\cite{Battitineco1992}.
\begin{equation}
    \label{eq:6}
    w_{t}= w_{t-1} - \eta {(H_a + \rho I)}^{-1} G^t
\end{equation}
Where $\rho > 0$  is the regularisation parameter, I is a identity matrix $\in$ $R^{d \times d}$ and $G^t$ = $\widehat{M_1}^t$. We use Sherman-Morrison formula of matrix inversion to directly compute the global Newton direction without forming and storing full regularized Hessian $(H_a + \rho I)$ and its inverse as shown below-\\

$(H_a+ \rho I)^{-1} G^t$= $(Z^t {V^t}^T+ \rho I)^{-1} G^t$\\

$(H_a+ \rho I)^{-1} G^t$=$\frac{G^t}{\rho}$ - $\frac{\frac{Z^t {V^t}^T G^t}{\rho^2}}{1+\frac{{V^t}^T Z^t}{\rho}}$\\  
where, $Z^t = \frac{V^t}{V^t[0]}$ and $V^t[0] = \frac{\partial^2 F}{\partial x_1^2}$ is the first element of $V^t$.\\

\begin{figure*}[!h]
  \centering
  \includegraphics[width=0.96\linewidth]{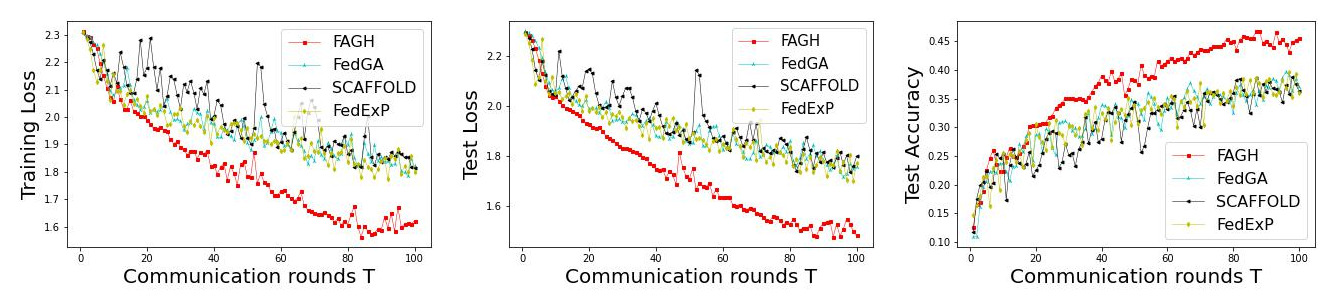}
  \caption{Comparisons of training loss, test loss and test accuracy on CIFAR10 image classification using LeNet5}\label{fig:p1}
\end{figure*}
\begin{figure*}[!h]
  \centering
  \includegraphics[width=0.96\linewidth]{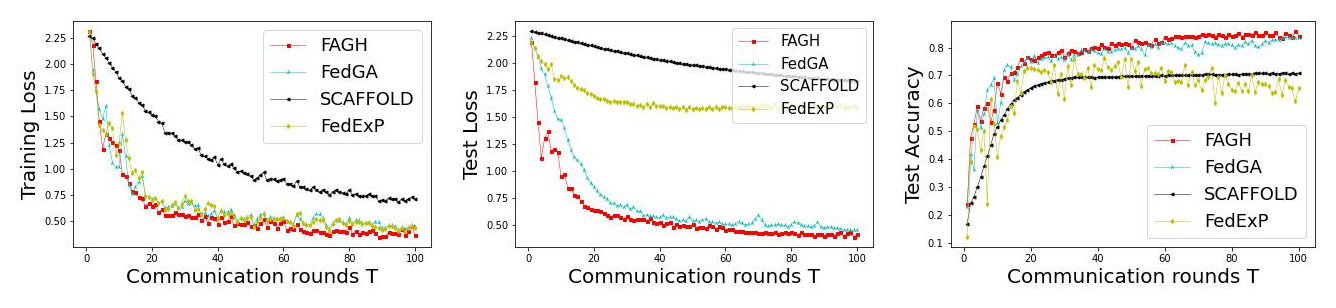}
  \caption{Comparisons of training loss, test loss and test accuracy on FashionMNIST image classification using CNN}\label{fig:p2}
\end{figure*}

\begin{figure*}[!h]
  \centering
  \includegraphics[width=0.96\linewidth]{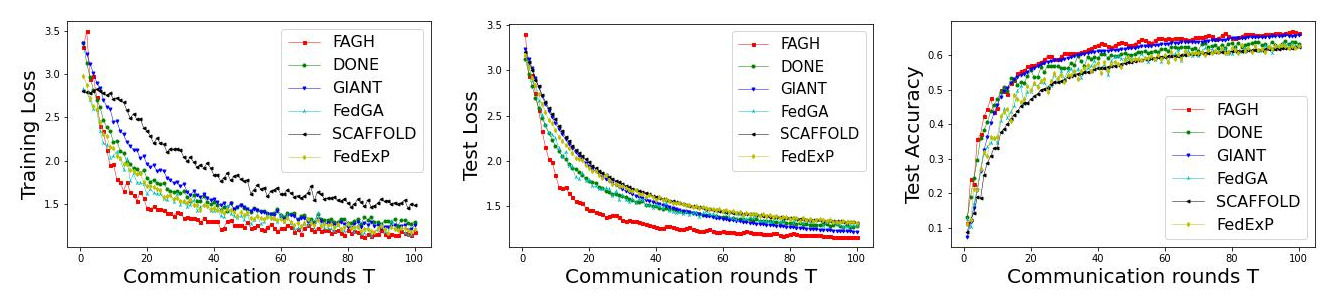}
  \caption{Comparisons of training loss, test loss and test accuracy on EMNIST  
 image classification using MLR}\label{fig:p3}
\end{figure*}

%%%%%%%%%%%%%%%%%%%%%%%%%%%%%%%%%%%%%%%%%%%%%%%%%%%%%%%%%%%
\begin{figure*}[!h]
  \centering
  \includegraphics[width=0.96\linewidth]{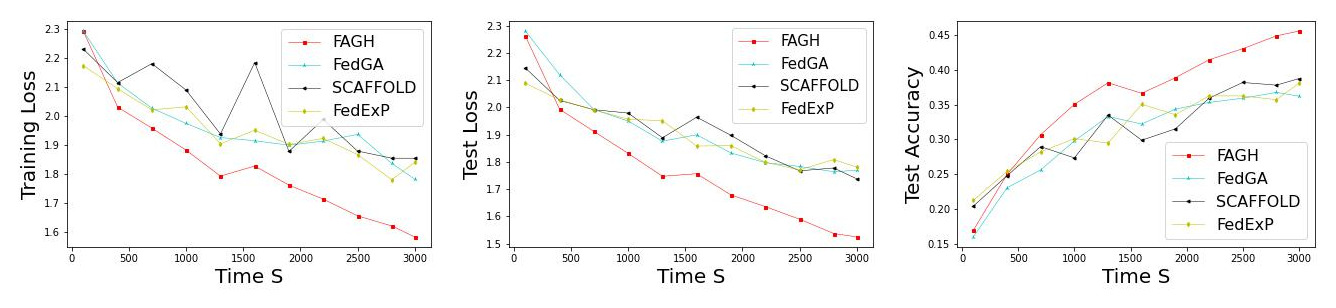}
  \caption{Time comparisons of training loss, test loss and test accuracy on CIFAR10 image classification using LeNet5}\label{fig:p4}
\end{figure*}
\begin{figure*}[!h]
  \centering
  \includegraphics[width=0.96\linewidth]{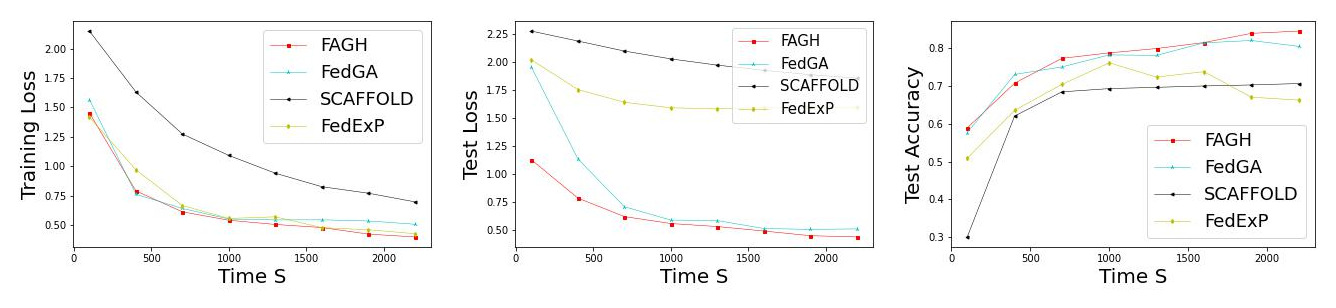}
  \caption{Time comparisons of training loss, test loss and test accuracy on FashionMNIST image classification using CNN}\label{fig:p5}
\end{figure*}

\begin{figure*}[!h]
  \centering
  \includegraphics[width=0.96\linewidth]{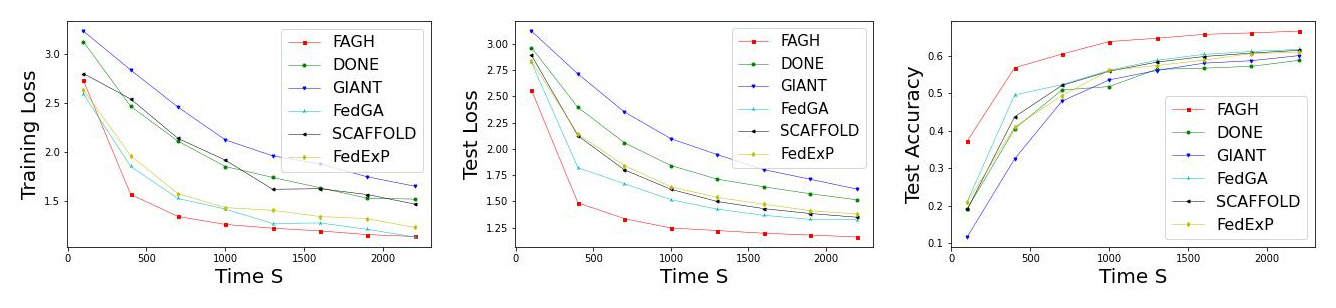}
  \caption{Time comparisons of training loss, test loss and test accuracy on EMNIST 
 image classification using MLR}\label{fig:p6}
\end{figure*}
\subsection{Complexities}

As FAGH is associated with the computation of gradient and the first row of the true Hessian in the local client, the local time and space complexities of FAGH are both O(d + d), which is similar to existing first-order based methods like SCAFFOLD \cite{Karimireddyicml2020}, FedDC \cite{Gaocvf2022}, etc. Second-order based methods like GIANT and DONE have a local complexity of O(md+d), which is higher than FAGH (Here $m > 1$ is the number of iterations associated with approximating the Newton update). The local space complexities of both DONE and GIANT are O( d + d), which is similar to FAGH. FedNL and Basis Matters store the previous step's Hessian matrix to approximate the current step's Hessian in local clients, which requires $O(d^2)$ local space complexity. Compared to this, FAGH is highly beneficial for resource-constrained local clients. In FAGH, as the server needs to retain information on the previous global gradient and the first row of the previous global Hessian to find the exponential moving averages of the current gradient and the current Hessian's first row, which results in the requirement of O(2d + 2d) space complexity on the server. As the server directly computes the global Newton direction with the help of the Sherman-Morrison formula for matrix inversion, the overall time complexity of the server is O(d). The server's space complexity of FAGH is only three times more than FedAvg, which is very less compared to FedSSO's server space complexity $O(d^2)$. Sustaining O(4d) space complexity may not be a big issue with the host server. One issue with FAGH is discovered that, like SCAFFOLD, FAGH is associated with O(2d) client-to-server communication cost, which can be handled using proper compression techniques before communication to the server. The applicability of compression in FL can be seen from FedNL and Basis Matters, where the local Hessian is compressed before sending it to the server.

\subsection{Applicability of FAGH}
As the approximation of the Hessian in section 5.1 is tailored for twice-differentiable and non-linear loss function F, FAGH is applicable only to optimization problems associated with such twice-differentiable and non-linear loss functions. For example, optimizing neural networks or multinomial logistic regression (MLR) with cross-entropy loss function entails dealing with such twice-differentiable and non-linear loss function. To assess the applicability of FAGH, we conducted extensive experiments on federated image classification tasks using machine learning and deep learning models with cross-entropy loss function, from which we observed promising outcomes from FAGH. 
%%%%%%%%%%%%%%%%%%%%%%%%%%%%%%%%%%%%%%%%%%%%
\section{Experimental setup}
To validate our proposed method, we conduct extensive experiments on heterogeneously partitioned CIFAR10 \cite{krizhevsky2009learning}, FashionMNIST \cite{xiao2017fashion} and EMNIST-letters \cite{cohen2017emnist} datasets. 
CIFAR10 comprises color images ($3 \times 32 \times 32$) of 10 classes with total 60000 samples (50000 training samples and 10000 test sample). FashionMNIST comprises grayscale images ($1 \times 28 \times 28$) of 10 classes with total 60000 samples (50000 training samples and 10000 test sample). EMNIST-letters comprises grayscale images ($1 \times 28 \times 28$ pixels) of handwritten uppercase and lowercase letters, divided into 26 classes. EMNIST-letters includes a total of 145,600 samples, with 124,800 for training and 20,800 for testing. To create heterogeneous data partitions for CIFAR10 and  FashionMNIST datasets, we use the same Dirichlet distribution based heterogeneous and unbalanced partition strategy as mentioned in the papers of \citeauthor{YurochkinAGGHK19} and \citeauthor{iclrWangYSPK20}. We simulate $P_i \sim Dir_K(0.2)$ and find a heterogeneous partition by allocating a $P_{(i,j)}$ proportion of the samples of $i^{th}$ class to $j^{th}$ client. As we use very small value of Dirichlet distribution's concentration parameter (0.2), each client may not get samples of all the classes, which indicates a high degree of data heterogeneity across all the clients. For our experiments, we use K=200 clients. For EMNIST dataset, we utilizes similar partitioning strategy used in the paper of  \citeauthor{McMahanaistats2017}, where the data has been sorted with the class label and then distributed. We create 400 shards of size 312 and assign 2 shards to each of the 200 clients. 

For CIFAR10 image classification, we use LeNet5 model \cite{lecun2015lenet}. For FashionMNIST, we use a custom convolutional neural network (CNN) model (total 1475338 trainable parameters) with two convolutional layers and three fully connected layers. After each convolutional layer, we use batch normalization, ReLU activation and max-pooling. After first fully connected layer, we use a dropout of 0.25. For EMNIST-letters, we use multinomial logistic regression (MLR) model. For all the federated image classification tasks, we use crossentropy loss function. 

We compare our algorithm with existing state-of-the-art federated learning algorithms such as SCAFFOLD, FedGA, FedExP, GIANT and DONE. To consider partial device participation in FL, we use $40\%$ of total client's participation in each communication round. We do extensive experiments with a wide set of hyper-parameters for all the methods and find the best performing model for each method by considering minimum training $\&$ test losses and maximum test accuracy. We use FedGA $\beta$, FAGH $\rho$ and FedExP $\epsilon$  $\in \{1, 0.5, 0.1, 0.01, 0.001\}$, learning rate $\eta \in \{1, 0.5, 0.1, 0.01, 0.001, 0.0001\}$, FAGH $\beta_1 = 0.9$ $\&$ $\beta_2 = 0.99$, number of Rechardson iterations for DONE=10, number of CG iterations for GIANT = 10 and $\alpha_{DONE} \in  \{0.01, 0.05\}$. For FedExP, we use SGD with momentum (0.9) optimizer for finding local updates. 
We use total number of communication rounds T=100. We implement all the methods using Tesla V100 GPU and PyTorch1.12.1+cu102. we use seed=0 and batch size = 512. For each dataset, we use same initialization and same settings for all the existing and proposed methods.

We compare our algorithm with DONE and GIANT in MLR based classification task. We tried to compare our algorithm with DONE and GIANT in CNN based classification tasks . But unfortunately, we did not find suitable hyperparameters for DONE and GIANT for this CNN based implementation. This may be due to their assumption of strongly convex loss function.
\begin{table}[h]
    \centering
    \begin{tabular}{lllll}
        \hline
        Method  & $30\%$ & $35\%$ & $40\%$ & $45\%$ \\
        \hline
        \textbf{FAGH}  & 18 & 29 & 43 & 78 \\
        FedGA   & 23 & 54 & 96 & ... \\
        SCAFFOLD  & 38 & 66 & ... & ... \\
        FedExP & 24 & 48 & 97 & ... \\
        \hline
    \end{tabular}
    \caption{Comparisons of communication rounds' while achieving differnet test accuracies results on CIFAR10 with LeNet5 model. Here, "..." means that the algorithm can not be able to achieve the accuracy within FL iterations T = 100.}
    \label{tab1}
\end{table}

\begin{table}[h]
    \centering
    \begin{tabular}{lllll}
        \hline
        Method  & $60\%$ & $70\%$ & $80\%$ & $85\%$ \\
        \hline
        \textbf{FAGH}   & 7 & 14 & 36 & 71 \\
        FedGA   & 7 & 12 & 46 & ... \\
        SCAFFOLD  & 13 & 35 & ... & ... \\
        FedExP & 8 & 18 & ... & ... \\
        \hline
    \end{tabular}
    \caption{Comparisons of communication rounds' while achieving differnet test accuracies results on FashionMNIST with CNN model. Here, "..." means that the algorithm can not be able to achieve the accuracy within FL iterations T = 100.}
    \label{tab2}
\end{table}

\begin{table}[h]
    \centering
    \begin{tabular}{lllll}
        \hline
        Method  & $40\%$ & $50\%$ & $60\%$ & $65\%$ \\
        \hline
        \textbf{FAGH}   & 6 & 11 & 25 & 58 \\
        FedGA   & 11 & 21 & 46 & ... \\
        SCAFFOLD  & 13 & 23 & 61 & ... \\
        FedExP & 11 & 19 & 55 & ... \\
        DONE & 7 & 12 & 34 & 96 \\
        GIANT & 8 & 12 & 33 & 79 \\
        \hline
    \end{tabular}
    \caption{Comparisons of communication rounds' while achieving differnet test accuracies results on EMNIST with MLR model. Here, "..." means that the algorithm can not be able to achieve the accuracy within FL iterations T = 100.}
    \label{tab3}
\end{table}

\subsection{Results}
Our experimental results are shown in figs. [\ref{fig:p1}, \ref{fig:p2}, \ref{fig:p3}, \ref{fig:p4}, \ref{fig:p5}, \ref{fig:p6}] and tables [\ref{tab1}, \ref{tab2}, \ref{tab3}]. From these figures, it may be observed that FAGH can decrease the train and test losses in less time and less communication rounds as compared to SCAFFOLD, FedGA, FedExP, GIANT $\&$ DONE. It also may be observed that FAGH can achieve better test accuracy at different time steps and communication rounds as compared to SCAFFOLD, FedGA, FedExP, GIANT $\&$ DONE. From the tables, it may be observed that FAGH takes comparatively less number of communication rounds for achieving different targeted test accuracies. As we use the same initialization and same settings for all the methods, we may claim that FAGH can provide faster FL training while achieving a certain precision of the global model performance in heterogeneous FL settings with partial clients participation. FAGH is easy to implement, as it has only two active tuning hyper-parameters, one is Hessian regularization parameters ($\rho$) and another one is learning rate ($\eta$). From our experiments, we noticed that same as ADAM \cite{KingmaB14}, we can standardize the exponential decay rates for the moment estimates  of FAGH ($\beta_1$ and $\beta_2$ ) to 0.9 and 0.99 respectively. 

%%%%%%%%%%%%%%%%%%%%%%%%%%%%%%%%%%%%%%%%%%%%%

\section{Conclusions}
We proposed a new Newton optimization-based FL training method, namely FAGH, by making use of the approximated global Hessian for accelerating the convergence of global model training in FL, thereby resolving the challenge of the heavy communication overhead in FL due to a large amount of communication rounds needed to train the global model toward convergence. FAGH is beneficial for practical implementation in terms of both local and server space complexities in comparison to existing Newton-based FL training algorithms. Experimental results demonstrate that FAGH outperforms several state-of-the-art FL training methods, including SCAFFOLD, FedGA, FedExP, GIANT, and DONE, in terms of the number of communication rounds and the time required to train the global model in FL to achieve the pre-specified performance objectives. In the future, we plan to investigate how to identify a set of local clients for participating in training the global model in an adaptive and privacy-preserving manner, e.g., by leveraging learning vector quantization \cite{qin2005initialization} and graph matching \cite{gong2016discrete} techniques, to further improve the convergence of the global model while keeping its performance in other aspects.

\bibliographystyle{named}
\bibliography{ijcai23}

\begin{thebibliography}{}

\bibitem[\protect\citeauthoryear{Agarwal \bgroup \em et al.\egroup }{2017}]{Agarwaljmlr2017}
Naman Agarwal, Brian Bullins, and Elad Hazan.
\newblock Second-order stochastic optimization for machine learning in linear time.
\newblock {\em J. Mach. Learn. Res.}, 18:116:1--116:40, 2017.

\bibitem[\protect\citeauthoryear{Battiti}{1992}]{Battitineco1992}
Roberto Battiti.
\newblock First and second-order methods for learning: Between steepest descent and newton's method.
\newblock {\em Neural Comput.}, 4(2):141--166, 1992.

\bibitem[\protect\citeauthoryear{Bischoff \bgroup \em et al.\egroup }{2021}]{bischoff2021second}
Sebastian Bischoff, Stephan G{\"u}nnemann, Martin Jaggi, and Sebastian~U Stich.
\newblock On second-order optimization methods for federated learning.
\newblock {\em arXiv preprint arXiv:2109.02388}, 2021.

\bibitem[\protect\citeauthoryear{Cohen \bgroup \em et al.\egroup }{2017}]{cohen2017emnist}
Gregory Cohen, Saeed Afshar, Jonathan Tapson, and Andre Van~Schaik.
\newblock Emnist: Extending mnist to handwritten letters.
\newblock In {\em 2017 international joint conference on neural networks (IJCNN)}, pages 2921--2926. IEEE, 2017.

\bibitem[\protect\citeauthoryear{Dandi \bgroup \em et al.\egroup }{2022}]{DandiBJgradalign2022}
Yatin Dandi, Luis Barba, and Martin Jaggi.
\newblock Implicit gradient alignment in distributed and federated learning.
\newblock In {\em Thirty-Sixth {AAAI} Conference on Artificial Intelligence, {AAAI} 2022, Thirty-Fourth Conference on Innovative Applications of Artificial Intelligence, {IAAI} 2022, The Twelveth Symposium on Educational Advances in Artificial Intelligence, {EAAI} 2022 Virtual Event, February 22 - March 1, 2022}, pages 6454--6462. {AAAI} Press, 2022.

\bibitem[\protect\citeauthoryear{Derezinski and Mahoney}{2019}]{DDerezinskiM19}
Michal Derezinski and Michael~W. Mahoney.
\newblock Distributed estimation of the inverse hessian by determinantal averaging.
\newblock In Hanna~M. Wallach, Hugo Larochelle, Alina Beygelzimer, Florence d'Alch{\'{e}}{-}Buc, Emily~B. Fox, and Roman Garnett, editors, {\em Advances in Neural Information Processing Systems 32: Annual Conference on Neural Information Processing Systems 2019, NeurIPS 2019, December 8-14, 2019, Vancouver, BC, Canada}, pages 11401--11411, 2019.

\bibitem[\protect\citeauthoryear{Dinh \bgroup \em et al.\egroup }{2020}]{DinhTN20NeurIPS}
Canh~T. Dinh, Nguyen~Hoang Tran, and Tuan~Dung Nguyen.
\newblock Personalized federated learning with moreau envelopes.
\newblock In Hugo Larochelle, Marc'Aurelio Ranzato, Raia Hadsell, Maria{-}Florina Balcan, and Hsuan{-}Tien Lin, editors, {\em Advances in Neural Information Processing Systems 33: Annual Conference on Neural Information Processing Systems 2020, NeurIPS 2020, December 6-12, 2020, virtual}, 2020.

\bibitem[\protect\citeauthoryear{Dinh \bgroup \em et al.\egroup }{2022}]{Dinhieeetrans2022}
Canh~T. Dinh, Nguyen~H. Tran, Tuan~Dung Nguyen, Wei Bao, Amir~Rezaei Balef, Bing~Bing Zhou, and Albert~Y. Zomaya.
\newblock {DONE:} distributed approximate newton-type method for federated edge learning.
\newblock {\em {IEEE} Trans. Parallel Distributed Syst.}, 33(11):2648--2660, 2022.

\bibitem[\protect\citeauthoryear{Gao \bgroup \em et al.\egroup }{2022}]{Gaocvf2022}
Liang Gao, Huazhu Fu, Li~Li, Yingwen Chen, Ming Xu, and Cheng{-}Zhong Xu.
\newblock Feddc: Federated learning with non-iid data via local drift decoupling and correction.
\newblock In {\em {IEEE/CVF} Conference on Computer Vision and Pattern Recognition, {CVPR} 2022, New Orleans, LA, USA, June 18-24, 2022}, pages 10102--10111. {IEEE}, 2022.

\bibitem[\protect\citeauthoryear{Gong \bgroup \em et al.\egroup }{2016}]{gong2016discrete}
Maoguo Gong, Yue Wu, Qing Cai, Wenping Ma, A.~K. Qin, Zhenkun Wang, and Licheng Jiao.
\newblock Discrete particle swarm optimization for high-order graph matching.
\newblock {\em Information Sciences}, 328:158--171, 2016.

\bibitem[\protect\citeauthoryear{Jhunjhunwala \bgroup \em et al.\egroup }{2023}]{fedexp}
Divyansh Jhunjhunwala, Shiqiang Wang, and Gauri Joshi.
\newblock Fedexp: Speeding up federated averaging via extrapolation.
\newblock {\em CoRR}, abs/2301.09604, 2023.

\bibitem[\protect\citeauthoryear{Karimireddy \bgroup \em et al.\egroup }{2020}]{Karimireddyicml2020}
Sai~Praneeth Karimireddy, Satyen Kale, Mehryar Mohri, Sashank~J. Reddi, Sebastian~U. Stich, and Ananda~Theertha Suresh.
\newblock {SCAFFOLD:} stochastic controlled averaging for federated learning.
\newblock In {\em Proceedings of the 37th International Conference on Machine Learning, {ICML} 2020, 13-18 July 2020, Virtual Event}, volume 119 of {\em Proceedings of Machine Learning Research}, pages 5132--5143. {PMLR}, 2020.

\bibitem[\protect\citeauthoryear{Ketkar and Ketkar}{2017}]{ketkar2017stochastic}
Nikhil Ketkar and Nikhil Ketkar.
\newblock Stochastic gradient descent.
\newblock {\em Deep learning with Python: A hands-on introduction}, pages 113--132, 2017.

\bibitem[\protect\citeauthoryear{Kingma and Ba}{2015}]{KingmaB14}
Diederik~P. Kingma and Jimmy Ba.
\newblock Adam: {A} method for stochastic optimization.
\newblock In Yoshua Bengio and Yann LeCun, editors, {\em 3rd International Conference on Learning Representations, {ICLR} 2015, San Diego, CA, USA, May 7-9, 2015, Conference Track Proceedings}, 2015.

\bibitem[\protect\citeauthoryear{Krizhevsky \bgroup \em et al.\egroup }{2009}]{krizhevsky2009learning}
Alex Krizhevsky, Geoffrey Hinton, et~al.
\newblock Learning multiple layers of features from tiny images.
\newblock 2009.

\bibitem[\protect\citeauthoryear{LeCun and others}{2015}]{lecun2015lenet}
Yann LeCun et~al.
\newblock Lenet-5, convolutional neural networks.
\newblock {\em URL: http://yann. lecun. com/exdb/lenet}, 20(5):14, 2015.

\bibitem[\protect\citeauthoryear{Li \bgroup \em et al.\egroup }{2020a}]{Licorr2020}
Tian Li, Anit~Kumar Sahu, Manzil Zaheer, Maziar Sanjabi, Ameet Talwalkar, and Virginia Smith.
\newblock Feddane: {A} federated newton-type method.
\newblock {\em CoRR}, abs/2001.01920, 2020.

\bibitem[\protect\citeauthoryear{Li \bgroup \em et al.\egroup }{2020b}]{Limlsys2020}
Tian Li, Anit~Kumar Sahu, Manzil Zaheer, Maziar Sanjabi, Ameet Talwalkar, and Virginia Smith.
\newblock Federated optimization in heterogeneous networks.
\newblock In {\em Proceedings of Machine Learning and Systems 2020, MLSys 2020, Austin, TX, USA, March 2-4, 2020}. mlsys.org, 2020.

\bibitem[\protect\citeauthoryear{Li \bgroup \em et al.\egroup }{2020c}]{Liiclr2020}
Xiang Li, Kaixuan Huang, Wenhao Yang, Shusen Wang, and Zhihua Zhang.
\newblock On the convergence of fedavg on non-iid data.
\newblock In {\em 8th International Conference on Learning Representations, {ICLR} 2020, Addis Ababa, Ethiopia, April 26-30, 2020}, 2020.

\bibitem[\protect\citeauthoryear{Li \bgroup \em et al.\egroup }{2021}]{Licvpr2021}
Qinbin Li, Bingsheng He, and Dawn Song.
\newblock Model-contrastive federated learning.
\newblock In {\em {IEEE} Conference on Computer Vision and Pattern Recognition, {CVPR} 2021, virtual, June 19-25, 2021}, pages 10713--10722, 2021.

\bibitem[\protect\citeauthoryear{Liu and Nocedal}{1989}]{Liu1989mp}
Dong~C. Liu and Jorge Nocedal.
\newblock On the limited memory {BFGS} method for large scale optimization.
\newblock {\em Math. Program.}, 45(1-3):503--528, 1989.

\bibitem[\protect\citeauthoryear{Ma \bgroup \em et al.\egroup }{2022}]{Macorr2022}
Xin Ma, Renyi Bao, Jinpeng Jiang, Yang Liu, Arthur Jiang, Jun Yan, Xin Liu, and Zhisong Pan.
\newblock Fedsso: {A} federated server-side second-order optimization algorithm.
\newblock {\em CoRR}, abs/2206.09576, 2022.

\bibitem[\protect\citeauthoryear{Martens and Grosse}{2015}]{Martens2015jmlr}
James Martens and Roger~B. Grosse.
\newblock Optimizing neural networks with kronecker-factored approximate curvature.
\newblock In Francis~R. Bach and David~M. Blei, editors, {\em Proceedings of the 32nd International Conference on Machine Learning, {ICML} 2015, Lille, France, 6-11 July 2015}, volume~37 of {\em {JMLR} Workshop and Conference Proceedings}, pages 2408--2417. JMLR.org, 2015.

\bibitem[\protect\citeauthoryear{McMahan \bgroup \em et al.\egroup }{2017}]{McMahanaistats2017}
Brendan McMahan, Eider Moore, Daniel Ramage, Seth Hampson, and Blaise~Ag{\"{u}}era y~Arcas.
\newblock Communication-efficient learning of deep networks from decentralized data.
\newblock In {\em Proceedings of the 20th International Conference on Artificial Intelligence and Statistics, {AISTATS} 2017, 20-22 April 2017, Fort Lauderdale, FL, {USA}}, volume~54, pages 1273--1282. {PMLR}, 2017.

\bibitem[\protect\citeauthoryear{Nazareth}{2009}]{nazareth2009conjugate}
John~L Nazareth.
\newblock Conjugate gradient method.
\newblock {\em Wiley Interdisciplinary Reviews: Computational Statistics}, 1(3):348--353, 2009.

\bibitem[\protect\citeauthoryear{Qian \bgroup \em et al.\egroup }{2022}]{QianISR22}
Xun Qian, Rustem Islamov, Mher Safaryan, and Peter Richt{\'{a}}rik.
\newblock Basis matters: Better communication-efficient second order methods for federated learning.
\newblock In Gustau Camps{-}Valls, Francisco J.~R. Ruiz, and Isabel Valera, editors, {\em International Conference on Artificial Intelligence and Statistics, {AISTATS} 2022, 28-30 March 2022, Virtual Event}, volume 151 of {\em Proceedings of Machine Learning Research}, pages 680--720. {PMLR}, 2022.

\bibitem[\protect\citeauthoryear{Qin and Suganthan}{2005}]{qin2005initialization}
A.~K. Qin and P.~N. Suganthan.
\newblock Initialization insensitive {LVQ} algorithm based on cost-function adaptation.
\newblock {\em Pattern Recognition}, 38(5):773--776, 2005.

\bibitem[\protect\citeauthoryear{Safaryan \bgroup \em et al.\egroup }{2022}]{SafaryanIQR22}
Mher Safaryan, Rustem Islamov, Xun Qian, and Peter Richt{\'{a}}rik.
\newblock Fednl: Making newton-type methods applicable to federated learning.
\newblock In Kamalika Chaudhuri, Stefanie Jegelka, Le~Song, Csaba Szepesv{\'{a}}ri, Gang Niu, and Sivan Sabato, editors, {\em International Conference on Machine Learning, {ICML} 2022, 17-23 July 2022, Baltimore, Maryland, {USA}}, volume 162 of {\em Proceedings of Machine Learning Research}, pages 18959--19010. {PMLR}, 2022.

\bibitem[\protect\citeauthoryear{Shamir \bgroup \em et al.\egroup }{2014}]{Shamiricml2014}
Ohad Shamir, Nathan Srebro, and Tong Zhang.
\newblock Communication-efficient distributed optimization using an approximate newton-type method.
\newblock In {\em Proceedings of the 31th International Conference on Machine Learning, {ICML} 2014, Beijing, China, 21-26 June 2014}, volume~32, pages 1000--1008, 2014.

\bibitem[\protect\citeauthoryear{Swokowski}{1979}]{swokowski1979calculus}
Earl~William Swokowski.
\newblock {\em Calculus with analytic geometry}.
\newblock Taylor \& Francis, 1979.

\bibitem[\protect\citeauthoryear{Tan \bgroup \em et al.\egroup }{2021}]{Tancorr2021}
Alysa~Ziying Tan, Han Yu, Lizhen Cui, and Qiang Yang.
\newblock Towards personalized federated learning.
\newblock {\em CoRR}, abs/2103.00710, 2021.

\bibitem[\protect\citeauthoryear{Tankaria \bgroup \em et al.\egroup }{2021}]{Tankaria2021corr}
Hardik Tankaria, Dinesh Singh, and Makoto Yamada.
\newblock Nys-curve: Nystr{\"{o}}m-approximated curvature for stochastic optimization.
\newblock {\em CoRR}, abs/2110.08577, 2021.

\bibitem[\protect\citeauthoryear{Vuchkov}{2022}]{vuchkov2022hessian}
Radoslav~G Vuchkov.
\newblock {\em Hessian Approximations for Large-Scale Inverse Problems Governed By Partial Differential Equations}.
\newblock PhD thesis, UC Merced, 2022.

\bibitem[\protect\citeauthoryear{Wang \bgroup \em et al.\egroup }{2018}]{Wangneurips2018}
Shusen Wang, Farbod Roosta{-}Khorasani, Peng Xu, and Michael~W. Mahoney.
\newblock {GIANT:} globally improved approximate newton method for distributed optimization.
\newblock In {\em Advances in Neural Information Processing Systems 31: Annual Conference on Neural Information Processing Systems 2018, NeurIPS 2018, December 3-8, 2018, Montr{\'{e}}al, Canada}, pages 2338--2348, 2018.

\bibitem[\protect\citeauthoryear{Wang \bgroup \em et al.\egroup }{2020a}]{iclrWangYSPK20}
Hongyi Wang, Mikhail Yurochkin, Yuekai Sun, Dimitris~S. Papailiopoulos, and Yasaman Khazaeni.
\newblock Federated learning with matched averaging.
\newblock In {\em 8th International Conference on Learning Representations, {ICLR} 2020, Addis Ababa, Ethiopia, April 26-30, 2020}. OpenReview.net, 2020.

\bibitem[\protect\citeauthoryear{Wang \bgroup \em et al.\egroup }{2020b}]{Wangneurips2020}
Jianyu Wang, Qinghua Liu, Hao Liang, Gauri Joshi, and H.~Vincent Poor.
\newblock Tackling the objective inconsistency problem in heterogeneous federated optimization.
\newblock In {\em Advances in Neural Information Processing Systems 33: Annual Conference on Neural Information Processing Systems 2020, NeurIPS 2020, December 6-12, 2020, virtual}, 2020.

\bibitem[\protect\citeauthoryear{Xiao \bgroup \em et al.\egroup }{2017}]{xiao2017fashion}
Han Xiao, Kashif Rasul, and Roland Vollgraf.
\newblock Fashion-mnist: a novel image dataset for benchmarking machine learning algorithms.
\newblock {\em arXiv preprint arXiv:1708.07747}, 2017.

\bibitem[\protect\citeauthoryear{Yurochkin \bgroup \em et al.\egroup }{2019}]{YurochkinAGGHK19}
Mikhail Yurochkin, Mayank Agarwal, Soumya Ghosh, Kristjan~H. Greenewald, Trong~Nghia Hoang, and Yasaman Khazaeni.
\newblock Bayesian nonparametric federated learning of neural networks.
\newblock In Kamalika Chaudhuri and Ruslan Salakhutdinov, editors, {\em Proceedings of the 36th International Conference on Machine Learning, {ICML} 2019, 9-15 June 2019, Long Beach, California, {USA}}, volume~97 of {\em Proceedings of Machine Learning Research}, pages 7252--7261. {PMLR}, 2019.

\end{thebibliography}

\end{document}